\pdfoutput=1

\documentclass[11pt]{article}

\usepackage[]{emnlp2021}

\usepackage{times}
\usepackage{latexsym}
\usepackage{amsmath}
\usepackage{amsfonts}
\usepackage{amssymb}
\usepackage{multirow}

\definecolor{lightblue}{RGB}{135,206,250}
\newcommand{\blue}[1]{\colorbox{lime}{\textbf{#1}}}
\newcommand{\red}[1]{\colorbox{red}{\textbf{#1}}}

\usepackage[T1]{fontenc}

\usepackage[utf8]{inputenc}

\usepackage{microtype}

\usepackage{hyperref}
\usepackage{url}
\usepackage{soul}
\usepackage[normalem]{ulem}
\usepackage{wrapfig,lipsum,booktabs}
\usepackage{hyperref}
\usepackage{url}
\usepackage{dsfont}
\usepackage{booktabs} 
\usepackage{caption}
\usepackage{amsmath}
\usepackage{mathtools}
\usepackage{sidecap}
\usepackage{multirow}
\usepackage{epsfig,graphics,float}
\usepackage{float}
\usepackage{todonotes}
\usepackage{subfig,balance,lineno}
\usepackage{float}
\usepackage{verbatim}

\usepackage{wrapfig}

\title{Learning from Language Description: Low-shot Named Entity
Recognition via Decomposed Framework}

\author{Yaqing Wang$^{\S}$, Haoda Chu$^\dagger{}$, Chao Zhang$^\diamond$ and Jing Gao$^{\S}$ \\
  $^{\S}$Purdue University,
  $^\dagger{}$Microsoft,
  $^\diamond$Georgia Institute of Technology\\
  wang5075@purdue.edu, haochu@microsoft.com,\\ chaozhang@gatech.edu, jinggao@purdue.edu}

\begin{document}
\maketitle
\begin{abstract}
In this work, we study the problem of named entity recognition (NER)  in a low resource
scenario, focusing on few-shot and zero-shot settings. Built upon large-scale pre-trained language models, we propose a novel NER framework, namely SpanNER,  which learns from natural language supervision and enables the identification of never-seen entity classes without using in-domain labeled data. We perform  extensive experiments on 5 benchmark datasets and evaluate the proposed method in the few-shot learning, domain transfer and zero-shot learning settings. The experimental results show that the proposed method can bring  10\%, 23\% and 26\%  improvements in average over the best baselines in few-shot learning, domain transfer and zero-shot learning settings respectively.
\end{abstract}

\section{Introduction}

Named entity recognition (NER)  aims at identifying and categorizing spans of text into a pre-defined set of classes, such as people, organizations, and locations. As a fundamental language understanding task, NER is widely adopted in question answering~\citep{molla2006named}, information retrieval~\citep{guo2009named} and other language understanding applications~\citep{nadeau2007survey, ritter2012open, peng2020soloist}.  Recent advances with pre-trained language models like BERT~\citep{DBLP:conf/naacl/DevlinCLT19}, GPT-2~\citep{radford2019} and RoBERTa~\citep{DBLP:journals/corr/abs-1907-11692} have shown remarkable success in NER . However, the success of these large-scale models still relies on fine-tuning them on large amounts of in-domain labeled data. Unfortunately, obtaining NER annotations not only is expensive and time consuming, but also may not be feasible in many sensitive user applications due to data access and privacy constraints.  This motivates us to study the problem of low-shot NER.

Low-shot NER focuses on identifying custom entities in a new domain with only a few in-domain examples or even without any in-domain labeled data, which are refereed to as few-shot NER and zero-shot NER respectively.  The success of low-shot NER requires the model to be capable of transferring learned knowledge to recognize new entity classes.  Conventional NER models usually treat each class as a one-hot vector (represented by a class label) for training, and thus the trained model cannot capture the semantic meanings of those labels. In fact,  the trained model could be highly associated with known classes and it is difficult to transfer learned knowledge to novel entity 
classes.

\begin{figure*}[htb!]
 		\centering
	\includegraphics[width=0.8\textwidth]{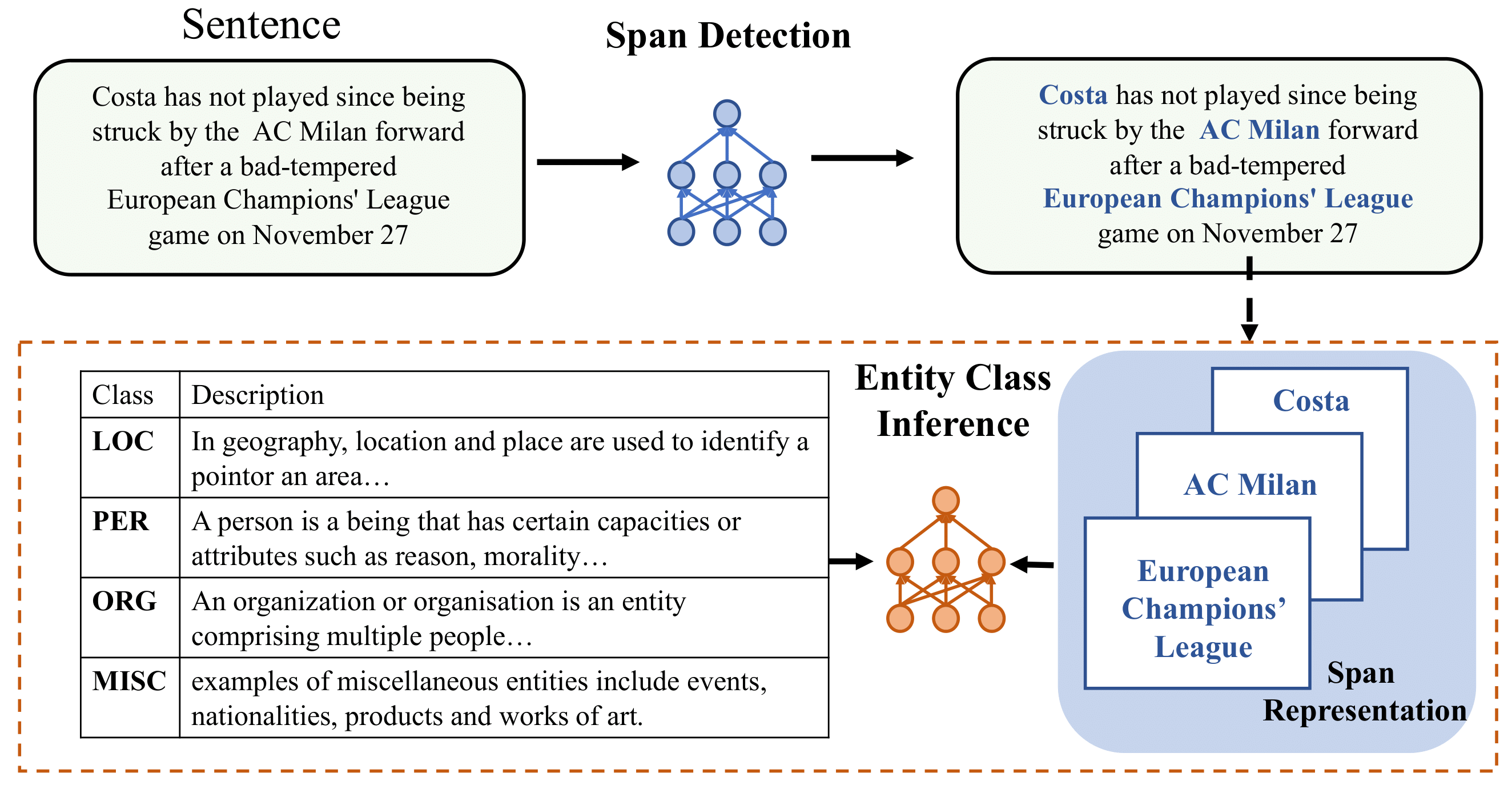}
 	  	\caption{An overview of the proposed NER system: SpanNER.}
 	 	 	\label{fig:framework}
 	 	 	\vspace{-0.1in}
\end{figure*}

To tackle this problem, several recent works~\citep{huang2020few, yang2020simple, hou2020few, Ziyadi2020ExampleBasedNE, wiseman2019label} employ prototype-based method to represent each class by a prototype based on the labeled examples and use nearest neighbor method for NER.
{However, each entity class in NER task may include several fine-grained entity classes and has diverse semantic meanings. Correspondingly, the tokens or entities belonging to the same entity class are not necessarily close to each other~\citep{huang2020few}, making it  challenging to represent each entity class by a prototype based on a few examples.} For example, \texttt{MISC}, one of the four entity classes in the benchmark dataset CoNLL03~\citep{Tjong2003Intro}, is a collection of {fine-grained entity classes} including events, nationalities, products and works of art. \texttt{FAC} is an entity class in the OntoNotes5~\citep{weischedel2012ontonotes} collection, including buildings, airports, highways, bridges and others.  {Thus, prototype-based methods may end up learning noisy representations of prototypes and cannot achieve satisfactory performance.}  Moreover, prototype-based methods have unavoidable reliance on labeled examples, thereby making them unable to extend to the zero-shot learning setting, which is also an important and practical scenario in the low-shot NER.

In this paper, we propose a simple yet effective method SpanNER which can tackle few-shot as well as zero-shot NER. Instead of deriving the representations of labels purely from a few labeled
examples, we propose to directly learn from the natural language descriptions of entity classes. {Such a choice provides a  flexible and precise way to obtain the semantic meanings of entity
 classes and enable zero-shot learning.}  Although using natural language as supervision has been explored in the context of zero-shot text classification, it is challenging to be adapted in the NER task. Unlike its use in text classification, natural language cannot provide direct supervision for token classification. { Inspired by machine reading comprehension (MRC) framework, we propose to decompose the NER task into two procedures: span detection and entity class inference, which can be jointly trained together.  However, it is challenging to employ MRC framework into NER task especially in the low-resource setting. The MRC framework usually needs a large amount of data for training, which is not available in the low resource scenario. To handle these challenges, we propose a class-agnostic span detection module which is equipped with token sampling mechanism to mitigate the imbalanced class issue and can be trained with limited labeled data. Moreover, to handle the challenging that {several fine-grained classes are included in one entity class}, we develop an entity class attention module to focus on the most relevant information in the given entity class description that corresponds to the extracted span.  Compared with direct adaption of MRC framework into NER~\cite{mrc}, the proposed method can bring more than 54\%, 30\% and 26\% improvement in average under few-shot learning, domain transfer and zero-shot learning settings respectively.} Figure~\ref{fig:framework} shows an overview of the proposed framework: the span detection stage aims to identify the span of text, and entity class inference is responsible for categorizing
extracted spans based on natural language description of pre-defined entity classes. We perform extensive experiments on 5 benchmark datasets and evaluate the proposed method in the few-shot learning, domain transfer and zero-shot learning settings. The experimental results show that the proposed method brings large improvements over the state-of-the-art methods across different settings.

{\noindent\textbf{Contributions.} Our model design is simple but distinguishes from that of the other NER works. To the best of our knowledge, we are the first one to learn entity class via natural language for NER task and the proposed method SpanNER achieves around 10\%, 23\% and 26\% improvement over state-of-the-art NER methods in few-shot learning, domain-transfer and zero-shot learning settings respectively.}

\section{Task Formulation}

NER is the process of locating and classifying named entities in text into predefined entity categories, such as person names, organizations, and locations.  Formally, given a sentence with $N$ tokens $X = \{x_1,..., x_N \}$, an entity or slot value draws from a span of tokens $s = [x_i,..., x_j] (0 \leq i \leq j \leq N)$ associated with an entity class $c \in \mathcal{C}$. The corresponding annotations for given sentence $X$ are denoted as $Y$.

\noindent \textbf{Few-shot NER} focuses on the NER task in a low-resource setting, where a system is only provided with a few in-domain labeled examples per entity class and the system needs to learn to identify custom entities in the domain. The task of $K$-shot NER refers to the setting with $K$ labeled input sentences per entity class $c \in \mathcal{C}$, and the training data can be denotes as $\mathcal{D}_\mathrm{train} = \{X_{i}, Y_{i}\}^{|\mathcal{C}|\times K}_{i=1}$. In this work, we leverage given training data $\mathcal{D}_\mathrm{train}$  for   model fine-tuning.

\noindent \textbf{Zero-shot NER} {focuses on a more challenging setting where a model is trained  with a set of entity classes and then tested on a dataset with a different set of entity classes.
Towards this end, zero-shot NER systems need to learn to generalize to unseen entity classes without using any labeled example.  The training data for zero-shot learning can be denoted as $\mathcal{D}_\mathrm{train}$ associated with an entity class set $\mathcal{C}_\mathrm{train}$, and the test data is denoted as $\mathcal{D}_\mathrm{test}$ associated with an entity class set $\mathcal{C}_\mathrm{test}$. Note that $\exists c_\mathrm{test} \in \mathcal{C}_\mathrm{test}$ but $c_\mathrm{test} \not\in \mathcal{C}_\mathrm{train} $}.


\section{Methodology}



In this paper, we study how to develop an effective model which can identify custom entities in a
novel domain with a small set of labeled data or even without using any labeled data. To this end,
we decompose NER task into two sub-tasks: span detection and entity class inference. The span
detection module is {class}-agnostic and can transfer knowledge across different entity classes. On top of span
detection, the entity class inference module takes extracted spans as input to infer the semantic
relationship between the spans with natural language description of entity classes. Learning from
natural language has an important advantage over existing categorical label learning methods, which is its ability to capture semantic meanings of labels and enable flexible zero-shot transfer.

\subsection{Span Detection}
\label{sec:span_detection}

Span detection is explored in the machine reading comprehension (MRC) frameworks~\citep{chen2017reading,seo2016bidirectional}, which predict the probability
for each token as the starting or ending of the answer span given a question. 
{However, it is challenging to directly adapt MRC framework for NER task especially in the low-resource setting.  First, for each entity class, the model needs to answer its associated natural language question and repeat this procedure until all the questions are answered~\cite{mrc}. Thus, such a method is not scalable when the number of entity classes increases and further exacerbates the imbalanced class issue compared to conventional NER framework. Second, the MRC framework usually needs a large amount of data, which is not available in the low resource scenario. {To handle these challenges, we propose a class-agnostic span detection module which  can share the knowledge across classes and develop a token sampling mechanism to mitigate imbalanced issue. The proposed span detection module takes input sequence as input without questions and can be trained with limited labeled data}.} Given an input sequence $X = \{x_1,..., x_N \}$, we first feed $X$ into a pre-trained  BERT~\citep{DBLP:conf/naacl/DevlinCLT19} to obtain token representations $\{\mathbf{x}_1,..., \mathbf{x}_N \} \in \mathbb{R}^{h \times N}$. Besides start and end index predictions, we also classify whether a token is a part of the entity span. For example, we get the score for each token being start as follows:
\begin{align}
s_\mathrm{start}(i) =  \mathbf{w}^{\intercal}_\mathrm{start} \cdot \mathbf{x}_{i}, 
\end{align}
where $\mathbf{w}_{\mathrm{start}} \in \mathbb{R}^{h \times 1}$ is the weight of the linear classifier. Correspondingly, the probability of a token being start index is:
\begin{align}
\small
p_\mathrm{start}(i) = \mathrm{sigmoid} (s_\mathrm{start}(i)).
\end{align}
The probability calculations for end and a part of a span  are the same as that of start index
prediction. We then compute the probability of a span $[i, j]$ being an  entity as:

\begin{eqnarray*}
\small
p_\mathrm{match}([i,j]) &=& \mathrm{sigmoid} \Big(s_\mathrm{start}(i) + s_\mathrm{end}(j) \\ &&+ \sum_{t=i}^j s_\mathrm{span}(t)\Big).
\end{eqnarray*}
The span detection loss consists of three parts: start prediction loss, end prediction loss and span matching loss. The loss function of start prediction can be represented as:
\begin{align}
\small
\mathcal{L}_\mathrm{start} = \frac{1}{N} \sum_{i=1}^N  \mathrm{CE} (p_\mathrm{start}(i), y_\mathrm{start}^i),
\end{align}
where $\mathrm{CE}$ represents cross-entropy function and $y_\mathrm{start}^i=1$ if token $x_i$ is the start of an entity.  The loss of end prediction can be calculated in a similar way.

\noindent\textbf{Mitigation of imbalanced class issue.} For an input sequence with length $N$, the number of span candidate is in a $N \times N$ scale, where most of them are negative span candidates. To mitigate the imbalanced class issue, we sample a subset of negative span candidates, denoted as $O^\mathrm{neg}$. The span candidate set which corresponds to gold spans is denoted as $O^\mathrm{pos}$. Instead of using all the negative spans, we propose to sample a subset of negative spans to mitigate imbalance class issue. More specifically, we set the sampling size  of  negative span candidates as $|O^\mathrm{neg}| = N - |O^\mathrm{pos}|$ to reach a  class ratio between positive and negative labels similar to the one in the start and end prediction losses. The role of such a mechanism is justified by an ablation study in Appendix. The span match loss is:
\begin{eqnarray*}
\small
  \mathcal{L}_\text{match} &=& -\frac{1}{N} \Big(\sum_{\substack{ (i,j) \in  O^\mathrm{pos}}}\log p_\mathrm{match}([i,j]) \\
    && + \sum_{\substack{ (i,j) \in  O^\mathrm{neg}}}\log{(1 - p_\mathrm{match}([i,j]))}\Big).
\end{eqnarray*}

The overall span objective consisting of three losses to be minimized is as follows:

\begin{align}
\mathcal{L}_\mathrm{span} = \mathcal{L}_\mathrm{start}  +  \mathcal{L}_\mathrm{end} +
\mathcal{L}_\mathrm{match}.
\end{align}

During inference, start and end indexes are first separately predicted. Then we select the consensus span between  match predictions and extracted (start, end) indexes to achieve final predictions.

\subsection{Natural Language Supervision}
\label{sec:adaptive}
Learning based on categorical labels only may  discard the semantic meanings of labels, and thus it is difficult to transfer knowledge from known classes to new entity classes. 
To mitigate this limitation, we propose to use natural language description\footnote{In this paper, we use the definitions of entity classes from Wikipedia or annotation guidelines as their language description.} of entity classes to provide supervision for entity class inference and enable zero-shot learning. However, different with zero-shot text classification or entity linking, the entity class description in NER may describe several fine-grained entity classes, making our setup more challenging.


\noindent \textbf{Mention Representation.}  Upon span detection, we can first obtain the mention span representation of each span candidate $[i, j]$ by averaging the embeddings of the span tokens. However,  entity class is usually a high level category including many entities. Thus, we need to add a linear transformation $\mathbf{w}_\mathrm{entity} \in \mathbb{R}^{h \times h}$ to project average span token embedding into the entity class space:
\begin{align}
\mathbf{e}_{i,j} = \mathbf{w}_\mathrm{entity} \cdot ( \frac{1}{j-i+1} \sum_{t=i}^j \mathbf{x}_t).
\end{align}
The description of an entity class $c$ is a sequence of tokens, denoted as $X^c=\{x^c_1, ..., x^c_K\}$ . In this paper, we  feed entity class description into another pre-trained BERT~\citep{bert}  to obtain its representations $\{\mathbf{x}^c_1, ..., \mathbf{x}^c_K\} \in \mathbb{R}^{h \times K}$ . Since there is limited data or even no data for training in the novel domain, we fixed the  parameters of this BERT to expedite transferring by  maintaining the embedding of entity class  description from source and novel domains in the same space.

\noindent \textbf{Entity Class Description Attention.} The entity class description may describe several fine-grained entity classes. To focus on the information in the description that corresponds to the extracted span, we propose to construct adaptive entity class representation. More specifically, we use multi-headed attention mechanism~\citep{Transformer}. Each single attention function can be described as mapping a query and a set of key-value pairs to an output. The query, key and value vector are denoted as $\mathbf{Q}$, $\mathbf{K}$ and $\mathbf{V}$ respectively. We use the aggregated mention vector $\mathbf{e}_{i,j} \in R^{ h \times 1}$ as query vector $\mathbf{Q}$ and use entity class description embedding $\mathbf{X}_c =[\mathbf{x}^c_{1},..., \mathbf{x}^c_{K}] \in \mathbb{R}^{ h \times K}$ as key vector  $\mathbf{K}$ and value vector $\mathbf{V}$. The output is computed as a weighted sum of the values, where the weight assigned to each value is computed by the dot-product function of the query with the corresponding key. Then multiple parallel attention heads can stabilize the learning mechanism. We represent the procedure of obtaining adaptive entity class representation $\mathbf{x}^c(e_{i,j}) \in \mathbb{R}^{ h \times 1}$ as:
\begin{align}
  \mathbf{x}^c(e_{i,j}) = \mathrm{MultiHead}(\mathbf{Q}, \mathbf{K}, \mathbf{V}).
\end{align}


The role of such a mechanism is empirically justified by the comparison between the proposed model and a reduced model (i.e., the proposed model without attention mechanism) in the experimental section.

\noindent \textbf{Entity Class Inference.} The entity class inference is to infer the relationship between entity class and extracted span.  We follow the zero-shot text classification~\citep{yin2019benchmarking} to cast this task into binary prediction: whether the extracted span belongs to given entity class or not.  The probability of the extracted span being in a given entity class $c$ is based on a  matching score between them:
\begin{align}
p (c | e_{i,j}) = \mathrm{sigmoid}( \mathbf{e}_{i j}^{\intercal} \mathbf{x}^c(e_{i,j})).
\end{align}

The loss for each extracted span $[i,j]$ is calculated as:
\begin{align}
\mathcal{L_\mathrm{entity}} ([i, j]) =  \frac{1}{|\mathcal{C}|} \sum_{c \in \mathcal{C}} \mathrm{CE}\Big(p (c | e_{i,j}), y\Big),
\end{align}
where $\mathcal{C}$ is a set of entity classes of interest, and $y$ is binary label which equals to $1$ when extracted span belongs to entity class $c$  and $0$ otherwise. We use $\mathcal{L_\mathrm{entity}}$ to denote the entity class inference loss for all extracted spans.

\noindent \textbf{Final Loss.} We jointly train span detection
and entity class inference  modules by optimizing the sum of their losses.


\section{Experiments}
\vspace{-0.05in}
In this section, we empirically study and compare the proposed method with state-of-the-art methods in few-shot learning, domain transfer and zero-shot learning settings.

\subsection{Experimental Setup}

\noindent \textbf{Dataset}. We perform large-scale experiments with five different datasets\footnote{https://github.com/juand-r/entity-recognition-datasets} including Named Entity Recognition tasks and user utterances for task-oriented dialog systems as summarized in Table~\ref{tab:dataset}. (a)  { CoNLL03}~\citep{Tjong2003Intro} is a collection of news wire articles from the Reuters Corpus with 4 entity classes. (b) { OntoNotes5}~\citep{weischedel2012ontonotes} is in general domain including 18 entity classes. (c)  { WNUT} 2017 ~\citep{derczynski2017results}  is collected from social media with 6 entity classes.  (d) MIT {Movie and Restaurant} corpus~\citep{Liu2013AsgardAP} consist of  user utterances for movie and restaurant domains with 12 and 8  classes.

\begin{table}[htb!]
\centering
    	\resizebox{\columnwidth}{!}{%
    \begin{tabular}{lcccc}
        \toprule
         { Dataset} & Domain & { \# Classes} & { \# Train} & { \# Test}  \tabularnewline\midrule
              CoNLL03 & News &  4 & 14K & 3.6K \tabularnewline
            OntoNotes5 & General & 18 & 60K & 8.3K  \tabularnewline
            WNUT & Social Media & 6 & 3.4K & 1.6K \tabularnewline
         Movie & Moive& 12 & 8.8K & 2.4K   \tabularnewline
         Restaurant & Restaurant &8 & 6.9K & 1.5K \tabularnewline
         \bottomrule
    \end{tabular}
    }
     \caption{Dataset summary.}
    \label{tab:dataset}

\end{table}

\noindent \textbf{Backbone} We use the the pre-trained  BERT$_\mathrm{base}$ uncased model ($\sim$110M parameters) as the backbone network. The inputs during training and inference are lowercased to make them case-insensitive. The implementation details and hyper-parameter configurations are presented in Appendix.

\subsection{Few-shot Learning}

\textbf{Setting.} In this subsection, we study how the proposed method performs in a few-shot supervision setting,  For 5-shot setting, we sample 5 sentences for each entity class from the training set and fine-tune models with sampled sentences. The experiment is repeated for 10 times to report the average F1 score. We also explore the role of distantly supervised learning in few-shot learning and the corresponding details are in Appendix.

\noindent \textbf{Baselines} The first baseline we use is a fully supervised BERT model trained on
all available training data {(3.4K-60K sentences)} which provides the ceiling performance for every task. Each of the other
models are trained on $5$ training sentences per class. We compare our method with BERT (same
backbone with ours) with {Beginning-Intermediate-Outside (BIO)} tagging mechanism as a comparison to evaluate the proposed model design
besides the backbone choice. {LC} and {Prototype} are abbreviations for linear classifier and
prototype-based methods from  {a recent few-shot NER work}~\citep{huang2020few}.  They use                 pre-trained model RoBERTa-base as their backbone model. MRC-NER~\citep{mrc}
casts NER task into machine reading comprehension and achieves the state-of-the-art performance on
several benchmark datasets. To study the role of attention mechanism proposed in
subsection~\ref{sec:adaptive}, we propose a reduced model {SpanNER-NoAttn}, which uses average operation instead of attention to aggregate entity class description.


\begin{table*}[htb]
\centering

\resizebox{0.9\linewidth}{!}{%
			\begin{tabular}{llllllll}
				\toprule
				 \textbf{Method} & \textbf{CoNLL03} &  \textbf{OntoNotes5} &  \textbf{WNUT} &     \textbf{Movie} & \textbf{Restaurant} & \textbf{Average}\\
			\midrule
							\multicolumn{6}{l}{\textbf{Full-supervision}}\\
				BERT & 91.1 & 87.8 & 47.1  & 87.9 & 79.0 &  78.6 \\
				\midrule

				 \multicolumn{6}{l}{\textbf{5-shot supervision}}\\
					BERT & 61.6&60.1&21.2&61.9&48.6 & 50.7  \\

					LC$^\dagger$  & 53.5  &57.7 &25.7&51.3&48.7 & 47.4 \\

					Prototype$^\dagger$  &58.5&53.3&\textbf{29.5}&38.0&44.1 & 44.7 \\
					MRC-NER & 28.5 & 49.8 &0.4&58.7 & 43.1&36.1\\

			 SpanNER-NoAttn (ours)  & \underline{68.4} (0.5)  &\underline{65.1} (0.3)&22.8 (0.4)& \underline{64.8} (0.3)& \underline{48.9} (0.2) & \underline{54.0}\\
						SpanNER (ours) &  \textbf{71.1} (0.4) &  \textbf{67.3} (0.5) & \underline{25.8} (0.3)$^*$ & \textbf{65.4} (0.4) & \textbf{49.1} (0.2) & \textbf{55.7} [$\uparrow$9.9\%]\\




				\bottomrule
			\end{tabular}

	}
		   \caption{F1 score comparison of models on different datasets. All models (except LC and Prototype) use the same BERT backbone. $^\dagger$ indicates results from \citep{huang2020few}. The highest scores are \textbf{bolded}, while the second highest score is \underline{underlined}. F1 score of our model for each task is followed by standard deviation and percentage improvement [$\uparrow$] is over the best baseline. $^*$Roberta is pre-trained on reddit dataset which is similar to WNUT. We change backbone from BERT to Roberta (same with Prototype's) and F1 of SpanNER on WNUT is \textbf{31.5} (0.3).}
	\vspace{-0.1in}
\label{tab:5_shot}
\end{table*}

\noindent \textbf{Performance}  We report the results of 5-shot supervision and distantly supervising pre-training plus 5-shot supervision in Table~\ref{tab:5_shot}. In the 5-shot supervision setting, we can observe that our methods outperform baseline BERT consistently, which shows the advantage of the proposed model design {in addition to} the benefits from backbone. The baseline Prototype leverages given support examples to conduct NER task and achieves lower performance compared with LC with the same backbone according to average F1. The reason may lie in that the tokens belonging to the same entity class are not necessarily close to each other~\citep{huang2020few}. Prototype achieves better performance on WNUT compared to SpanNER since Prototype is based on Roberta which is pre-trained on social media dataset reddit. We change backbone of the proposed model SpanNER from BERT to Roberta-bas and observe that SpanNER achieves 31.5 in term of F1 score and outperforms Prototype. The MRC-NER framework is a reading comprehension framework whose success relies on training on large-scale data and thus cannot achieve satisfactory performance in a few-shot setting. Overall, we observe that our methods largely outperform all methods including the models with the same BERT encoder as ours across different datasets.  The average performance improvement over the best baseline BERT is around 10\%. Moreover,  the comparison between SpanNER and SpanNER-NoAttn demonstrates that the improvement brought by attention mechanism is around 3.1\%.

\begin{figure}[htb!]
\vspace{-0.1in}
 		\centering	\includegraphics[width=0.7\linewidth]{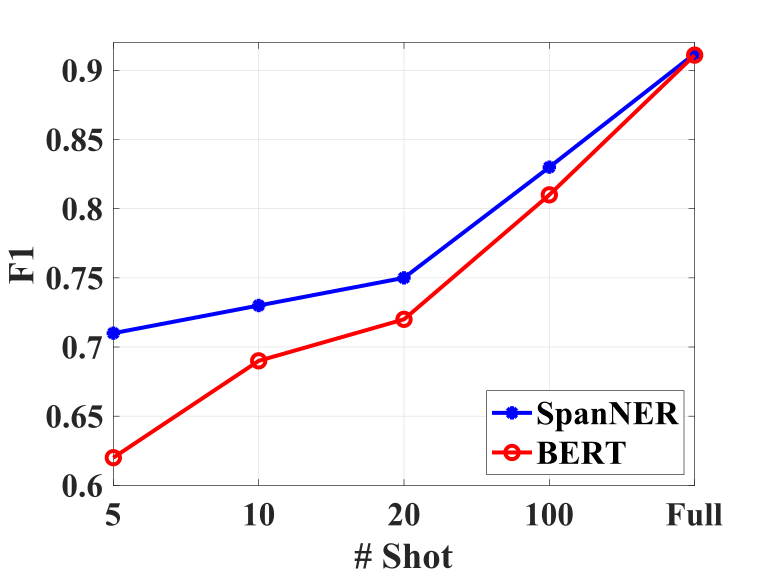}
 	  	\caption{Variation in model performance on varying shots on CoNLL03. ``Full'' indicates full supervision.}
 	 	 	\label{fig:shot_number}
 
\end{figure}

{\noindent\textbf{Varying the number of shots.} Table \ref{fig:shot_number} shows the improvement in the performance of SpanNER and BERT  when increasing the number of labels for each NER type in the CoNLL03 dataset. As we increase the amount of labeled training instances,  SpanNER  improves over BERT consistently.}

\subsection{Domain Transfer}
We evaluate the proposed model in another common scenario of adapting a NER model to a novel domain~\citep{yang2020simple}.  In this setting, we have a fully supervised source domain and a target domain with few-shot supervision. Following \citep{yang2020simple}, we use general domain (OntoNotes5) as a source domain and evaluate  models on News (CoNLL) and Social (WNUT) domains.




\begin{table} [htb]
\centering

\resizebox{\linewidth}{!}{%
\begin{tabular}{l|lll}
    \toprule
       \textbf{Models} & \textbf{CoNLL03} &  \textbf{WNUT} &  \textbf{Average}  \\ \midrule
       SimBERT $^\dagger$  & 28.6 &   7.7 & 18.2\\
       Prototypical Network $^\dagger$ & 65.9 &19.8 & 42.9\\
        PrototypicalNet+P\&D $^\dagger$  & 67.1 &  23.8 & 45.4 \\
        NNShot  $^\dagger$ & 74.3 & 23.9 & 49.1\\
       StructShot $^\dagger$   & 75.2 &  27.2 & 51.2 \\
       MRC-NER  & 64.1 & 32.6 &48.3\\
       	\midrule
         SpanNER-NoAttn (ours)& 80.1 (0.4) &42.0 (0.7) & 61.1\\
        SpanNER (ours) & \textbf{83.1} (0.5)&\textbf{43.1} (0.6) & \textbf{63.1} [$\uparrow$23.2\%]\\


    \bottomrule
\end{tabular}
}
\caption{F1 score comparison of models on CoNLL03 and WNUT datasets with 5-shot supervision  for domain transfer.   $^\dagger$ indicates results from~\citep{yang2020simple}.  }
\vspace{-0.1in}
\label{tab:domain_transfer}
\end{table}

\noindent \textbf{Baselines.} We adopt six state-of-the-art methods in the domain transfer setting as baselines. SimBERT is based on a pre-trained BERT encoder and the predictions are conducted by a nearest neighbor classifier ~\citep{yang2020simple}. 
Prototypical Network~\citep{snell2017prototypical} is a state-of-the-art few-shot classification system and is adopted by~\citep{fritzler2019few} for few-shot NER task. Upon Prototypical Network, PrototypicalNet+P\&D~\citep{hou2020few} adds pairwise embedding and dependency mechanism to gain further improvements.   StructShot and NNShot are proposed in \citep{yang2020simple}, which achieve state-of-the-art performance in this domain transfer setting. We include our reduced baseline SpanNER-NoAttn in this task as an ablation study and follow \citep{yang2020simple} to run our models five times and report average performance with standard deviation.

\noindent \textbf{Performance} Table~\ref{tab:domain_transfer} shows the results of baselines and the proposed methods on CoNLL03 and WNUT datasets. The proposed models achieve F1 scores 83.1 and 43.1  on CoNLL03 and WNUT respectively, bringing improvements over the best baseline 10.5\% and 32.2\% correspondingly. The proposed model effectively transfers learned knowledge to novel domains by learning from natural language descriptions instead of simple one-hot representation of entity classes.

\subsection{Zero-shot NER}
\noindent \textbf{Setting}  The zero-shot learning setting is motivated by the fact that new types of entities often emerge in some domains and sometimes the annotations in the target domain are not accessible.   Following zero-shot text classification setting~\citep{yin2019benchmarking}, we evaluate the proposed model in a common setting: \emph{label-partially-unseen}. In  label-partially-unseen setting,  a part of labels are unseen, enabling us to check the performance on unseen labels as well as seen labels.

\noindent \textbf{Baselines} Zero-shot NER task is rarely studied. The most state-of-the-art model for zero-shot NER is MRC-NER~\citep{mrc}, which conducts NER task by extracting answer spans given the questions of entity classes. Another baseline we use is the reduced model SpanNER-NoAttn. The comparison with this reduced model can demonstrate the role of entity class attention mechanism.

\begin{table}[htb]
\centering
\resizebox{\linewidth}{!}{%
			\begin{tabular}{lccccc}
				\toprule
				 \textbf{Method} &	\multicolumn{2}{c}{\textbf{CoNLL03}} &  & 	\multicolumn{2}{c}{\textbf{WNUT}}  \\
				 \cline{2-3}\cline{5-6}
				 & Overall & Unseen & & Overall & Unseen  \\
			\midrule
			 	MRC-NER & 39.1 & 14.5 &&24.0 & 7.4\\

			    SpanNER-NoAttn  & 39.0 &0.5&& 31.4 & 16.8  \\
				SpanNER  & \textbf{53.0} & \textbf{33.5} &&\textbf{35.4} & \textbf{18.8}\\
					\bottomrule
			\end{tabular}
		  }
		   \caption{F1 score comparison of models on CoNLL03 and WNUT datasets. {\em Overall} and {\em Unseen} indicate F1 scores of all entity classes and never-seen entity classes, respectively.  }
		 	\vspace{-0.1in}
\label{tab:0_shot}
\end{table}

\noindent \textbf{Performance} Table \ref{tab:0_shot} shows the F1 scores of MRC-NER and the proposed methods. MRC-NER based on reading comprehension framework is capable of conducting NER task  for never-seen classes. However, the span detection in MRC-NER is tightly coupled with {question} understanding, leading to more difficulty in handling unseen entity classes. In contrast, the proposed framework decomposes the NER task into span detection and entity class inference, avoiding the error propagation between two modules and thus delivering better performance. The comparison between SpanNER-NoAttn and SpanNER indicates the importance of attention mechanism in entity class description understanding, especially for never-seen entity classes.

\noindent \textbf{Entity Class Inference} We conduct experiments that disentangle entity class inference module from SpanNER so that the capability of this module can be evaluated. The span detection module cannot be separately evaluated because the span annotations on both datasets are associated with pre-defined entity classes.  To demonstrate the capability of entity class inference, we use gold spans to evaluate the performance of entity class inference. Table~\ref{tab:analysis} shows the performance of SpanNER-NoAttn and SpanNER. Comparing these two methods, we can observe that the attention mechanism helps improve the performance of the entity class inference.

\begin{table}[htb]
\centering

		\resizebox{0.9\columnwidth}{!}{
			\begin{tabular}{lcccc}
				\toprule
				 & \textbf{CoNLL03} &   \textbf{WNUT} \\
			\midrule

				\multicolumn{3}{l}{\textbf{Entity Class Inference}} & \\
				 SpanNER-NoAttn & 56.2 & 53.7\\
			    SpanNER & 60.2 & 57.0\\

			    \midrule
			    \multicolumn{3}{l}{\textbf{Annotation guidelines}} & \\
			    	SpanNER-NoAttn & 31.8  & 11.5 \\
			    SpanNER &  42.1 & 15.7 \\

				\bottomrule
			\end{tabular}
		  }
		   \caption{Experiments that demonstrate the performance of the entity class inference module and adopt annotating guidelines as entity class descriptions on CoNLL03 and WNUT datasets.}
		 	 
\label{tab:analysis}
\end{table}

\noindent \textbf{Class Description Construction} We set up experiments to study how the entity class description affects the model performance. In this experiment, we replace Wikipedia description of entity classes by annotation guidelines from CoNLL03 and WNUT datasets in the testing stage.  We can observe that the proposed models are still capable of identifying entities belonging to never-seen entity classes even though the descriptions in the testing stage are different from those in the training stage. The F1 scores drop compared to the scores when Wikipedia descriptions are used because training and test stages use different descriptions and the annotation guidelines do not include the semantic explanation of entity classes.

\noindent \textbf{Performance per Entity Class} We show F1 score per entity class on CoNLL03 and WNUT datasets in Table~\ref{tab:distribution}. We can observe the various degrees of recognizing different entity classes. First, the person names are easily recognized across different domains. The performance of person entity class  on WNUT is worse compared to that on CoNLL03, which may be due to the large domain shift in social media data. It is interesting to see that the performance of seen entity classes \texttt{LOC}, \texttt{location} and \texttt{product} is even worse than that of never-seen entity classes. To explain this interesting phenomenon, we provide a detailed analysis of error cases below.

\begin{table}[htb]
\centering
  
	\resizebox{0.9\columnwidth}{!}{
			\begin{tabular}{lrllr}
				\toprule
					\multicolumn{2}{c}{\textbf{CoNLL03}} && 	\multicolumn{2}{c}{\textbf{WNUT}} \\
					\cline{1-2}	\cline{4-5}
				 Entity Class& F1  &&  Entity Class& F1 \\
			\midrule

			PER&77.4&&	person & 59.1\\
			  ORG&58.5&&  creative-work$^*$& 19.3\\
			  MISC$^*$&33.5&&  corporation$^*$  & 19.1\\
			    LOC&5.9&& group$^*$&  18.3\\
			   -&-&&  location&14.4\\
			   -&-&&  product &11.0 \\

				\bottomrule
			\end{tabular}
			}
		
		   \caption{F1 score of SpanNER per entity class on CoNLL03 and WNUT datasets. $^*$ indicates unseen entity classes. }
		  \vspace{-0.1in}
\label{tab:distribution}
\end{table}

\noindent \textbf{Error Analysis} We manually examine the errors made by the proposed model on the CoNLL03 and WNUT test datasets and categorize these errors into 3 types. The error examples are presented in Appendix. (1) Different annotation guidelines on datasets. For instance, the description of \texttt{location} entity class in the source domain (OntoNotes5) is limited to mountain ranges and bodies of water, excluding countries, cities, states (these are included in the entity class \texttt{GPE}). Such a description is different from entity class \texttt{LOC} on CoNLL03 and \texttt{location} on WNUT.   (2) Domain shift. The domain shift leads to the difficulty in recognizing the entities belonging to seen entity classes.  (3) Description understanding. Description understanding is a crucial step for the success of zero-shot NER. For example,  \texttt{MISC} on CoNLL03 is a collection of diverse fine-grained entity classes including events, nationalities, products and works of art.

\section{Related Work}

The most related topics are few-shot and zero-shot NER, which are discussed as follows.

\noindent\textbf{Few-shot NER} aims to build a model that can recognize a new
class with a small number of labeled examples
quickly. Recent works~\citep{huang2020few, yang2020simple, hou2020few, Ziyadi2020ExampleBasedNE, wiseman2019label} exploit prototype-based methods to conduct NER tasks. 
Since tokens or entities belonging to the same entity class are not necessarily close to each other, prototype-based methods usually end up learning  noisy prototypes and may not achieve satisfactory performance. To further improve few-shot performance, \citep{hofer2018few, huang2020few} explores different pre-training strategies for few-shot NER, and \citep{wang2020adaptive, huang2020few} propose to leverage self-training to take advantage of additional unlabelled in-domain data. Although aforementioned few-shot NER works show the potential of additional data in improving performance of few-shot NER, they still suffer from the limitations of prototype or one-hot representations of labels in  transferring knowledge. Moreover, the aforementioned models cannot be applied in the zero-shot learning setting  due to either reliance on labeled support set or the adoption of one-hot label representation.

\noindent\textbf{Zero-shot NER} is to build a model that can recognize new classes without using corresponding labeled data. This setting is rarely studied in NER task. Zero-shot NER is important and practical in the real scenario since the annotations may not be accessible due to privacy and compliance restrictions for some sensitive user applications.  \citep{rei2018zero} has worked  on zero-shot sequence labeling task by using attention to infer binary token-level labels. However, their token level predictions are constrained to being binary and has to rely on sentence labels. These limitations prohibit the use of this method for NER task. MRC-NER~\citep{mrc} formulates NER task as a  machine reading comprehension task and enables zero-shot NER. However, the inference of MRC-NER for the single sentence needs to be conducted multiple times to collect results corresponding to all the entity types of interest, incurring expensive inference cost. Moreover, the reading comprehension framework needs to be trained with large-scale  dataset and is not effective in the few-shot setting. 
{\citep{logeswaran2019zero,wu2019scalable} propose to incorporate entity description for zero-shot entity-linking task. Nevertheless, they are different with ours in several aspects. First, the input to 
this work are sentences with pre-annotated entities while our task takes sentences  as  input and needs to conduct entity recognition and entity type inference jointly. Second, in the entity linking problem, the mentions in the sentences and Wikipedia descriptions refer to the same entities, while mentions in the sentences are only instances of entity classes in the NER task. Toward those challenges, SpanNER introduces how to capture class description which usually includes many fine-grained classes for a further improvement. }
Other zero-shot problems studied in NLP involve text classification~\citep{yin2019benchmarking}, entity typing~\citep{zhou2018zero}, word sense disambiguation~\cite{kumar2019zero} and relation extraction~\citep{levy2017zero}. These problems have different settings and challenges compared to zero-shot NER.

\vspace{-0.1in}
\section{Conclusions}
\vspace{-0.05in}
In this work, we study how to conduct NER in a low resource setting (when there are few or zero labeled data). To this end, we develop a novel NER framework SpanNER that decomposes NER task into span detection and entity class inference. This framework enables zero-shot NER and improves the performance of few-shot NER by capturing semantic meanings of entity classes.  Extensive experiments on 5 benchmark datasets and various settings demonstrate the effectiveness of the proposed model, particularly in the low-resource settings. 
\vspace{-0.1in}
\section*{Acknowledgment}
\vspace{-0.05in}
The authors would like to thank the anonymous referees for their valuable comments and helpful suggestions. This work is supported in part by the US National Science Foundation under grant NSF-IIS 1747614. Any opinions, findings, and conclusions or recommendations expressed in this material are those of the author(s) and do not necessarily reflect the views of the National Science Foundation.

\vspace{-0.2in}

\bibliography{anthology,custom}
\bibliographystyle{acl_natbib}

\clearpage
\appendix

\section{Implementation}

\noindent \textbf{Training}  We use AdamW~\cite{bert} as the optimizer. The learning rate is selected from $\{6e{-6}, 1e{-5}, 2e{-5}\}$ and batch size is selected from $\{8, 16\}$  according to validation set, coupled with a linear schedule with 1\% warmup. We clip
gradients to max norm $1.0$. For all training data sizes, we set the training epoch as 50. The max sequence length for input is 128 and the max sequence length for label description is 32. The dimension $h$ is 768, which is the embedding dimension of BERT. The pre-trained language model for label description is a BERT$_\mathrm{base}$ uncased model. For multi-head attention mechanism, the multi-head number is $4$, the hidden dimension size is $300$ and the Dropout probability is $0.2$.   The model is run on 4 NVIDIA Titan Xp GPU servers.

\begin{figure}[hbt!]
\vspace{-0.1in}
\subfloat[CoNLL03]{
\begin{minipage}{.49\linewidth}
\includegraphics[height=1.22in]{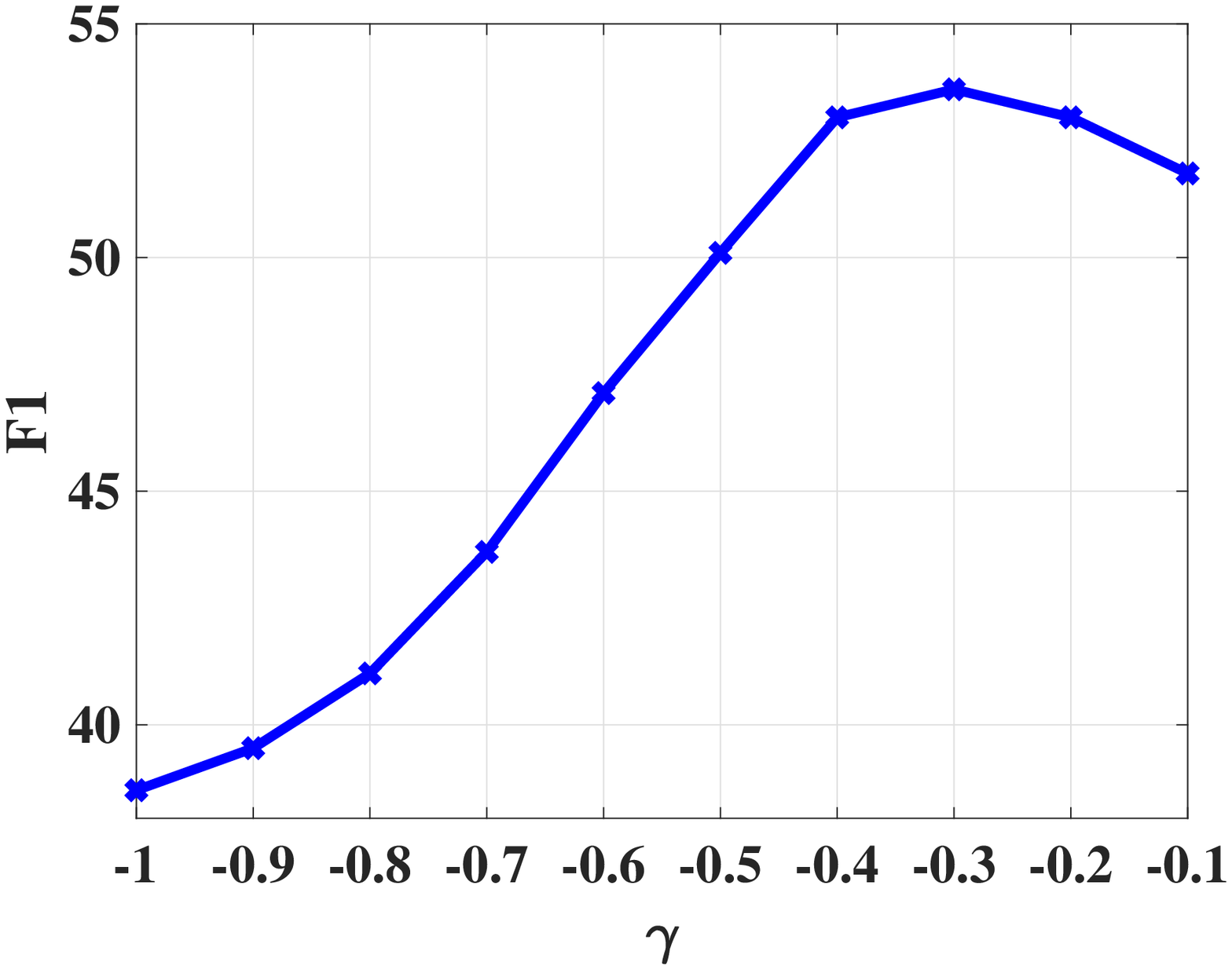}
\label{subfig:conll03}
\end{minipage}}
\subfloat[WNUT]{
\begin{minipage}{.49\linewidth}
\includegraphics[height=1.22in]{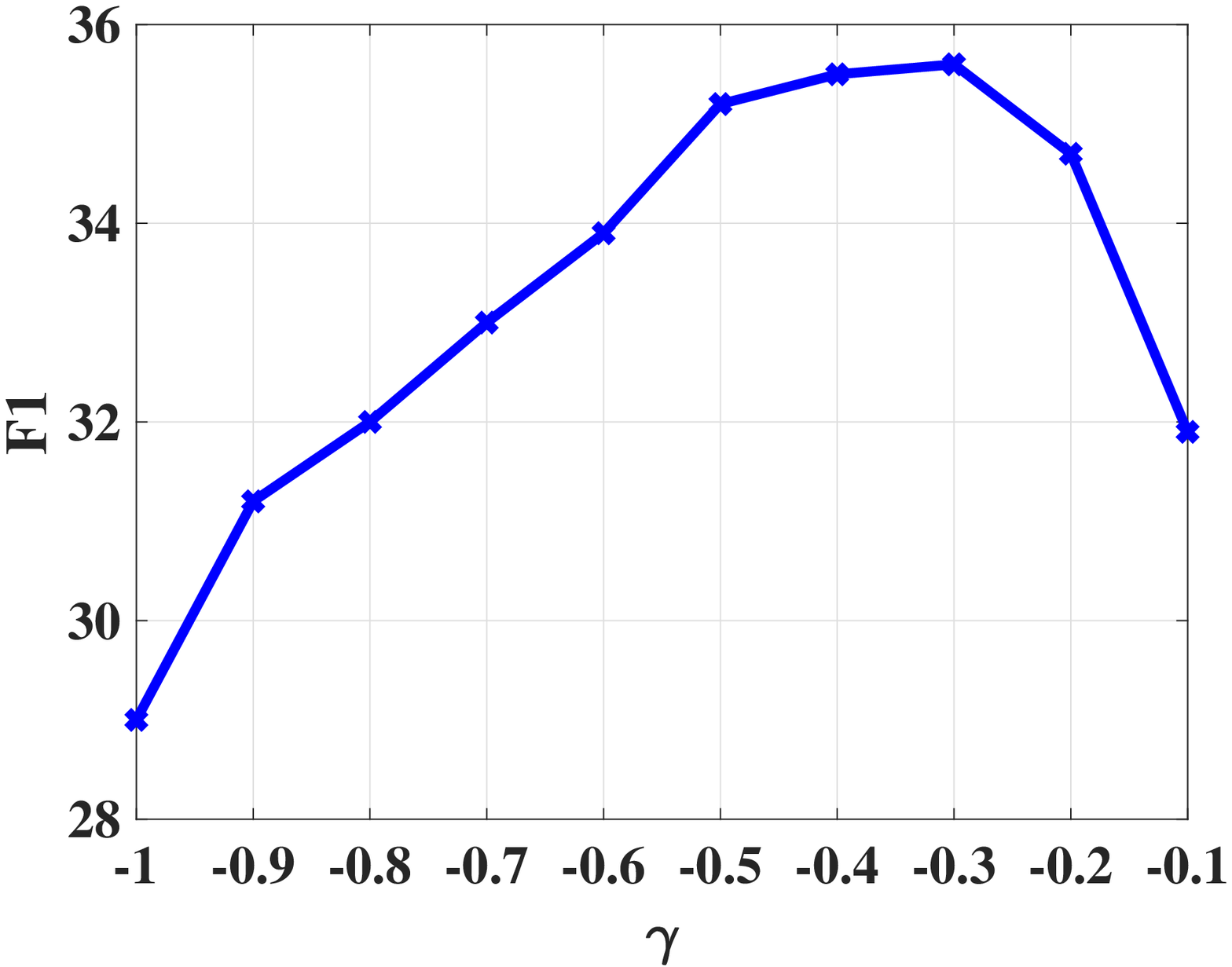}
\label{subfig:wnut17}
\end{minipage}}
 \caption{F1 score changes of our model w.r.t. varying threshold $\gamma$ values.}\label{fig:threshold}
\end{figure}

\noindent \textbf{Inference} For few-shot NER, the inference of span detection is introduced in Subsection~\ref{sec:span_detection}. Since the entity class inference is binary prediction for each entity class, we use $0.5$ as decision boundary to determine whether the extract span belongs to given entity class or not. During the zero-shot NER,  the mentions corresponding to never-seen classes may have lower detection scores compared to seen entity classes.  Thus, we calculate the entity inference score using softmax over all the entity classes to get a ``relative'' score. Since many extracted spans may not be associated with the entity classes of interest, we follow ~\cite{li2020efficient} to use a hyper-parameter threshold $\gamma$ to select identified entities based on the joint score of span detection and entity class inference  for span $[i, j]$. More specially, the joint score of span $[i ,j]$ can be calculated by $\log(p_\mathrm{match}([i,j])) +\log  p (c | e_{i,j})$. The thresholds of zero-shot are selected as $-0.4$ and $-0.5$ for CoNLL03 and WNUT respectively according to validation set.
 We investigate the effect of $\gamma$ in the Figure~\ref{fig:threshold}.

\noindent \textbf{Threshold $\gamma$} We show the F1 score changes of the proposed method on test datasets of CoNLL03 and WNUT with respect to varying $\gamma$ values in Figure~\ref{fig:threshold}. Similar trends can be observed on both datasets. The F1 scores first increase as threshold $\gamma$ values increase and then reach the peak in the middle points.

\section{Token Sampling}

Token sampling is introduced into the proposed model to mitigate the imbalance class issue, which degrade the efficiency and effectiveness of training especially in the low-resource scenario. To explore the role of token sampling, we conduct ablation study by removing the token sampling. The proposed baseline is denoted as SpanNER-NoSampling. We show the results  on CoNLL03 and OntoNoets5 with 5-shot supervision in Table.~\ref{tab:ablation_study}. Table.~\ref{tab:ablation_study} shows that the token sampling mechanism is effective and brings improvements around 4.7\% and 2.3\%  on CoNLL03 and OntoNotes5 respectively.

\begin{table}[htb]
\centering

		\resizebox{\columnwidth}{!}{
			\begin{tabular}{lll}
				\toprule
				 & \textbf{CoNLL03} &   \textbf{OntoNotes5} \\
			\midrule
	
				 SpanNER-NoSampling & 67.9 (0.2) & 65.8 (0.3)\\
			    SpanNER & \textbf{71.1} (0.4) & \textbf{67.3} (0.5)\\

				\bottomrule
			\end{tabular}
		  }
		   \caption{F1 score comparison of models on CoNLL03 and OntoNotes5 with 5-shot supervision. Overall highest scores are \textbf{bolded}. F1 score of models is followed by standard deviation.}
		 	   \vspace{-0.2in}
\label{tab:ablation_study}
\end{table}

\section{Distant Supervision}
\label{sec:distant_supervision}

\begin{table*}[htb]
\centering
	\label{tab:main_result}
\resizebox{0.9\linewidth}{!}{%
			\begin{tabular}{llllllll}
				\toprule
				 \textbf{Method} & \textbf{CoNLL03} &  \textbf{OntoNotes5} &  \textbf{WNUT} &     \textbf{Movie} & \textbf{Restaurant} & \textbf{Average}\\
			\midrule
							\multicolumn{6}{l}{\textbf{Full-supervision}}\\
				BERT & 91.1 & 87.8 & 47.1  & 87.9 & 79.0 &  78.6 \\
				\midrule

		\multicolumn{6}{l}{\textbf{Distantly Supervised Pretraining + 5-shot supervision}}\\

					BERT & 65.9 &64.9 &36.2&64.1&50.7 & 56.4\\

					LC$^\dagger$ & 61.4 &68.8&  34.2 &    53.1 & 49.1 & 53.3 \\
					Prototype$^\dagger$   &60.9 & 57.0 & 35.9 &  43.8 & 48.4 & 49.2  \\
					MRC-NER & 69.4 & 63.4 &36.3 & 64.6 & 48.7 & 56.5\\

				SpanNER-NoAttn (ours) & 74.4 (0.3)  &  66.5 (0.3) & 35.8 (0.6)  & 65.2 (0.2) & 49.3 (0.2) &58.3 \\
					SpanNER (ours) & 
					{\textbf{75.6}} (0.4)& 
					{\textbf{71.6}} (0.2) & 
					{\textbf{38.5}} (0.4) & {\textbf{67.8}} (0.3) & {\textbf{51.2}} (0.1) & {\textbf{60.9}}[$\uparrow$7.8\%]\\

				\bottomrule
			\end{tabular}

	}
		   \caption{F1 score comparison of models on different datasets. All models (except LC and Prototype) use the same BERT backbone. $^\dagger$ indicates results from \citep{huang2020few}. The overall highest scores are \textbf{bolded}. F1 score of our model for each task is followed by standard deviation. Percentage improvement [$\uparrow$] is over the best baseline in the corresponding setting.}
		   \vspace{-0.12in}
\label{tab:5_shot}
\end{table*}

We are also interested in the role of distantly supervised pre-training in the few-shot NER task, which is studied in a recent work~\citep{huang2020few}. We construct a Wikipedia distantly supervised dataset which contains 20 entity classes and around 1 million sentences. In this setting, we first pre-train NER models with distantly supervised data and then fine-tune the pre-trained models with 5-shot supervision.

To construct the training data, we use the May 2019 English Wikipedia dump and use the anchor text as the mention.  We select 20 popular entity types from  the fine-grained entity typing dataset Figer~\cite{ling2012fine} including
Locations, Organization, Person, City, Artist, Country, Author, Actor, Company, Event, Government, Sports team, Athlete, Title, Cemetery, Musician, Province, Building, Language, Politician. We retrieved entities belonging to those entity types by querying from Wikidata using SPARQL. Then we use string matching to assign entity types to anchor text.  We use a subset of the wikipedia data around 1M examples for training and 10K hold-out examples for validation.

The distantly supervised pre-training generally improves the performance for all the methods, confirming its potential in the low-resource NER task. Our model  achieves 60.9 in average F1 and  its average performance is improved by around 9.3\% compared to that in the setting of 5-shot supervision.  Overall, the proposed model  achieves the best performance on all the datasets and the improvement of SpanNER over the best baseline is around 7.6\% in average  terms of F1.

\section{Error Cases}

we show error examples made by our model in Table~\ref{tab:errors}. In the first example on CoNLL03, ``Japan'' is annotated as \texttt{LOC}. However, the entity class \texttt{location} on OntoNotes5 (source domain) does not include countries and thus SpanNER does not classify this span into \texttt{LOC}. We also observe some incorrect predictions  made by our model look reasonable. For instance, ``Outagmie County Circuit Court'' is identified as an entity belonging to \texttt{ORG}. Moreover, we find that our model correctly identifies ``Luebke'' as a \texttt{PER} entity which is not annotated in the dataset. On WNUT dataset, we find that many entities are difficult to identify especially that its context is very different with that of our source domain (OntoNotes5). For instance,  ``Watch What Else is Making News'' only includes limited context words ``Watch''.

\begin{table*}[bt]
\centering
\resizebox{0.85\linewidth}{!}{%
    \begin{tabular}{lp{14cm}}
    \toprule
    
        \textbf{Dataset} &  \textbf{Example Error} \\
         \midrule
         CoNLL03 & \red{Japan} then laid siege to the \red{Syrian} penalty area for most of the game but rarely  breached the \red{Syrian} defence. \\
         &Gold: \red{Japan} $\to$ LOC \\
         &Gold: \red{Syrian} $\to$  MISC\\
         &Gold:\red{Syrian}  $\to$ MISC \\
           CoNLL03 & {Japan} then laid siege to the \blue{Syrian} penalty area for most of the game but rarely  breached the \blue{Syrian} defence. \\
             &Pred: \blue{Syrian} $\to$ MISC \\
         &Pred:\blue{Syrian} $\to$  MISC \\

        \midrule
        
          CoNLL03 & In sentencing \red{Darrel Voeks}, 38, to a 10-year prison term on Thursday, \red{Outagmie County} Circuit Court Judge \red{Dennis Luebke} said he was ``a thief by habit."\\
          & Gold: \red{Darrel Voeks} $\to$ PER\\
          & Gold:\red{Outagmie County} $\to$ LOC\\
          & Gold: \red{Dennis Luebke} $\to$ PER \\
          CoNLL03 & In sentencing \blue{Darrel Voeks}, \blue{38}, to a 10-year prison term on Thursday, \blue{Outagmie County Circuit Court} Judge \blue{Dennis Luebke} said he was ``a thief by habit."\\ 
          & Pred: \blue{Darrel Voeks} $\to$ PER\\
          & Pred:\blue{38} $\to$ MISC\\
          & Pred: \blue{Outagmie County Circuit Court} $\to$ ORG \\
          & Pred: \blue{Dennis Luebke} $\to$ PER \\
           \midrule
          CoNLL03 & You are narcissitic, Luebke said at the sentencing, adding  \red{Voeks}  should pay restitution of more than \$100,000 to the farming family who had hired him. \\
          & Gold:  \red{Voeks} $\to$ PER\\
          CoNLL03 & You are narcissitic, \blue{Luebke} said at the sentencing, adding  \blue{Voeks}  should pay restitution of more than \$100,000 to the farming family who had hired him. \\
          & Pred: \blue{Luebke} $\to$ PER\\
          & Pred: \blue{Voeks} $\to$ PER\\
          
          \midrule
          CoNLL03 & 
          This cannot endure, ``\red{Marlow} told \red{BBC} television's \red{Newsnight} programme on Thursday." \\ 
           & Gold:  \red{Marlow} $\to$ PER\\
           & Gold: \red{BBC} $\to$ ORG\\
           & Gold: \red{Newsnight} $\to$ MISC \\
          CoNLL03 & 
          This cannot endure, ``\blue{Marlow} told \blue{BBC television's} Newsnight programme on Thursday." \\ 
           & Pred: \blue{Marlow} $\to$ PER\\
          & Pred: \blue{BBC television's} $\to$ ORG\\
           \midrule
          WNUT & Visuals of the avalanche site in \red{Gurez sector}.\\
          &Gold: \red{Gurez sector} $\to$ location\\
          
           WNUT & Visuals of the avalanche site in \blue{Gurez} sector.\\
           & Pred: \blue{Gurez} $\to$ location \\
           \midrule
           WNUT& Watch \red{What Else is Making News } \\
           & Gold: \red{What Else is Making News } $\to$ creative-work \\
           WNUT& Watch {What Else is Making News } \\
           \midrule
           WNUT& [Rip \red{Chad}]  (https://www.reddit.com/r/soccer/comments/5mi9bl/granada\_and\_ ...) \\
           &Gold: \red{Chad} $\to$ person \\
             WNUT& [\blue{Rip Chad}]  (https://www.reddit.com/r/soccer/comments/5mi9bl/granada\_and\_ ...) \\
              &Pred: \blue{Rip Chad} $\to$ person \\
             \midrule
             WNUT&Step 2: [\red{Google}] (http://www.google.co.nz) \\
             &Gold: \red{Google} $\to$ creative-work \\
              WNUT&Step 2: [\blue{Google}] (http://www.google.co.nz) \\
               &Gold: \blue{Google} $\to$ corporation \\

          \bottomrule
    \end{tabular}

}
    \caption{Error examples made by our model SpanNER in zero-shot NER setting. \red{} denotes gold spans and \blue{} denotes predicted spans.}
    \label{tab:errors}
\end{table*}

\end{document}